%% file: main.tex
\documentclass[runningheads]{llncs}
\pdfoutput=1
\usepackage{graphicx}

\usepackage{amsmath,amssymb,amsfonts}

\usepackage{nameref}
\usepackage{hyperref}

\usepackage{etoolbox}
\appto\UrlBreaks{\do\-}

\newcommand{\q}[1]{{``\textit{#1}''}}

\usepackage{xcolor}

\begin{document}

\title{Automating Speedrun Routing:\\
Overview and Vision}
%
\author{Matthias Gro\ss\inst{1}\orcidID{0000-0003-2711-6938} \and
Dietlind Z\"uhlke\inst{2}\orcidID{0000-0003-3751-5887} \and
Boris Naujoks\inst{2}\orcidID{0000-0002-8969-4795}}
\authorrunning{M. Gro\ss~ et al.}
\institute{Advanced Media Institute, TH K\"oln, Germany \\ \email{matthias.gross2@th-koeln.de} \and
Institute for Data Science, Engineering, and Analytics, TH K\"oln, Germany
\email{\{dietlind.zuehlke,boris.naujoks\}@th-koeln.de}
}
\maketitle              

\input{content/abstract}

\input{content/introduction}

\input{content/categories}

\input{content/relatedWork}

\input{content/approaches}

\input{content/algorithm}

\input{content/conclusions}

\input{bibliography/ludography}

%
%
%

\bibliographystyle{splncs04}

\bibliography{bibliography/references}

\end{document}

%% file: content/abstract.tex
\begin{abstract}

    Speedrunning in general means to play a video game fast,
    i.e. using all means at one's disposal to achieve a given goal in the least amount of time possible.
    To do so, a speedrun must be planned in advance, or routed, as referred to by the community.
    This paper focuses on discovering challenges and defining models needed when trying to approach the problem of routing algorithmically.
    To do so, this paper is split in two parts.
    The first part provides an overview of relevant speedrunning literature, extracting vital information and formulating criticism.
    Important categorizations are pointed out and a nomenclature is built to support professional discussion.
    The second part of this paper then refers to the actual speedrun routing optimization problem.
    Different concepts of graph representations are presented and their potential is discussed.
    Visions both for problem modeling as well as solving are presented and assessed regarding suitability and expected challenges.
    Finally, a first assessment of the applicability of existing optimization methods to the defined problem is made,
    including metaheuristics/EA and Deep Learning methods.

\keywords{speedrun \and routing \and graph models \and metaheuristics}

\end{abstract}

%% file: content/introduction.tex
\section{Introduction}
\label{sec:introduction}

What remains after you first time finished your favorite game? There are different options like changing to another game, play it once again in a different way or getting deeper into the game and try to solve it faster. And then once again, even faster. This approach ultimately leads to speedrunning, i.e. trying to solve the game as fast as possible. That has become a rapidly growing sub-community surrounding video games. One might consider special techniques, later referred to as glitches, to speed up the run or strictly object to these. Either way, for a lot of players, speedrunning presents the ultimate challenge.

The charity driven speedrunning live event \emph{Awesome Games Done Quick 2020} has attracted a maximum of 237,523 concurrent viewers \cite{GDQStatus}
and from 2011 to 2020 the event's donation total has been raised from 52,519.83 to 3,164,002.06 USD \cite{GDQTrackerEvent}.
Despite its popularity, scientific research on this specific field of esports is still only found sporadically.
In one of the first works that formally covered speedrunning, Newman describes it as
\q{[...] concerned with completing videogames in as speedy a time as possible} \cite{NewmanPlayingVideogames08}.

To do so, speedrunners traverse the progression mechanics of a given game as quickly as possible.
This can mean very different things depending on the game, from going around a racing track optimally to
solving puzzles in the most efficient order.

Being a discipline building heavily on optimization, it's only natural that speedrunning has become subject of various approaches
to enhance this optimization algorithmically.
As diverse as the game mechanics are, so are the means by which they have been tried to be supported. 
Finding successful policies for finishing a game is often approached using (Deep) Reinforcement Learning (see e.g. \cite{Szita2012}). 
For speedrunners, however, policies from Reinforcement Learning are a too detailed level of strategy.
They prefer to fix only the way points between helpful events and leave the rest to the skill of the player.
Thus we will focus on a different approach that allows optimizing the order of certain events and way points in the game's progression.

During the course of this work, Nintendo's famous Action-Adventure title \emph{The Legend of Zelda: Ocarina of Time} (\emph{OoT}) will be used as a working example.

The first part of this paper introduces speedrunning in general and builds understanding of the associated \emph{routing} problem, the main focus of this paper.
In \emph{\nameref{sec:categories}} (\ref{sec:categories}),
this work elaborates on the general description of speedrunning as well as the terminology as used in the community.
The \emph{\nameref{sec:relatedWork}} section (\ref{sec:relatedWork}) focuses on existing scholarly works
regarding speedrunning, including nomenclature to support professional discussions.
Connections between these works are made and criticism is formulated at the end of the section.

Equipped with this knowledge, the second part of this paper approaches the actual speedrun routing optimization problem.
The \emph{\nameref{sec:approaches}} (\ref{sec:approaches}) section presents approaches and ideas to formalize the problem.
Different models are envisioned and emerging challenges are critically discussed.
After these approaches to define the underlying problem, the \emph{\nameref{sec:algorithm}} (\ref{sec:algorithm}) section
elaborates on possible approaches to design algorithms for this problem,
also taking metaheuristics / EA and Deep Learning methods into account.
Finally, the paper closes with \emph{\nameref{sec:conclusions}} (\ref{sec:conclusions}).
All game titles used are listed in the \emph{\nameref{sec:ludography}} at the end of this paper.

%% file: content/categories.tex
\section{Speedruns and Categories}
\label{sec:categories}

Speedrunning in general means to play a game fast,
i.e. using all means at one's disposal to achieve a given goal in the least amount of time possible.
However, what is at one's disposal and what constitutes the given goal can vary widely.
To be comparable to each other and to give consistency for the runners, there are rules imposed on speedruns.
These rules are decided upon by the game's community, often by means of polls.
For a run to be listed on its leaderboard, it must comply to these rules, proven by uploaded footage of the corresponding runs. 
As the largest accumulation of speedrun footage and leaderboards, the website \href{https://www.speedrun.com/}{speedrun.com} will be used as point of reference here \cite{speedruncom}.
Some of the rules concern the form of proof each runner must submit with a claimed completion time,
but this paper will only take the rules imposed on actual gameplay into account.
There are some site-wide rules that apply to all runs on speedrun.com, e.g. the ban of any hardware manipulation.
There can also be game-wide rules all runs of this particular game must comply to.
Most of the site's leaderboards additionally list multiple \emph{categories} for the same game.
These categories are defined by different rulesets for a game's speedruns, each with their own leaderboard.
Although multiple games might share similar \emph{categories}, this does not necessarily mean any similarity beyond the names.
The underlying rulesets can differ largely from game to game.
Popular categories for many games are \emph{any\%}, \emph{100\%} and \emph{Glitchless}.

\begin{description}
\item[any\%]

Runs under the category \emph{any\%} usually don't have any additional rules imposed on them,
other than getting from a defined starting state to a defined ending state by any means possible, while respecting community-wide rules.
The name is derived from the fact, that any percentage of the game might be completed before ending the run.
This also usually means that these speedruns involve a lot of unconventional gameplay
using programming mistakes and inconsistencies in the game to lower the completion time as much as possible.
This unrestricted use of such \emph{glitches} and \emph{exploits} 
-- both of which will be discussed in more detail later on in \emph{\nameref{sec:relatedWork}} (\ref{sec:relatedWork}) --
often change the gameplay to a degree that makes the run look very different from a regular playthrough of the game.

\item[100\%] runs need to accomplish every objective in the game before completing it.
The definition of ``every objective in the game'' has to be agreed upon by the community and defined in the category's ruleset.
If the game provides a completion percentage display, e.g. in a saving, loading or status menu, it is used as the authoritative reference most of the time.
The nature of these runs vary heavily depending on the game being run.
As there often are a lot of objectives to achieve, the usage of glitches is prevalent.

\item[Glitchless]
Sometimes, newfound glitches completely trivialize a game's speedrun or parts of it.
Other times, the game's community finds a speedrun without glitches more appealing to do or watch \cite{Scully-BlakerRecuratingAccidentSpeedrunning16}. 
In such cases, the glitches in question can be banned in the categories' rulesets, or \emph{Glitchless} leaderboards can be created. 
As the name suggests, \emph{Glitchless} categories prohibit the use of \emph{glitches}, leading to runs more similar to casual playthroughs.
Again, what constitutes a \emph{glitch} is agreed upon by the community, as it is not a simple distinction \cite{RicksandTwereWellIt21}.
Alternatively, the ruleset can list all allowed techniques, all banned techniques, or both. 
\end{description}

Categories can be combined as applicable,
e.g. \emph{OoT} has, among others, an \emph{any\%}, a \emph{100\%}, a \emph{Glitchless any\%} and a \emph{Glitchless 100\%} leaderboard on speedrun.com.
The resulting difference in permitted techniques and completion requirements have a heavy impact on the run.

Another important categorization of speedruns is the differentiation between a \emph{Tool Assisted Speedrun} (TAS) and a \emph{Real-Time Attack} (RTA) run.
RTA runs are performed by a human in real-time.
Although the term ``speedrunning'' can be seen as more of an umbrella term,
RTA runs are the most prevalent form of speedrunning and are generally tied to the term unless specified. 
A TAS, in contrast, is prerecorded and assisted by tools, such as being able to create exact game state save points at arbitrary times.
Every input for every frame or simulation step in the game is handcrafted and scripted beforehand 
and then played back to create a perfectly optimized speedrun.
TASs are often created to demonstrate what could be theoretically achieved in a perfect run, or just for entertainment.
A variation of TASs are \emph{Low Optimization Tool Assisted Demonstrations} (LOTADs as referred to in the speedrunning community),
which usually demonstrate more exotic ways of playing a game or are used as reference for other runs.

One challenge that every speedrun, regardless of category, has to accommodate is \emph{routing}.
With regard to speedrunning, routing is the act of planning out a playthrough of a game in a fashion that needs the least amount of time possible.
Often, the game is broken down to individual events, depending on the game
(locations or characters to visit, items to collect, checkpoints to drive by etc.).
Then, a route trough the game world has to be constructed, that covers all events needed for game completion.
In many cases, specific vital events have preconditions in form of other events, that have to be incorporated into the route.
A good route is crucial for optimizing any given speedrun.
That said, this statement of course holds only true as long as the player has any choice on the ordering of the events.
Perfectly linear games may not have a strong focus on routing, if any at all.

%% file: content/relatedWork.tex
\section{Related Work}
\label{sec:relatedWork}

Aside from the categories and nomenclature presented earlier, there are also some scholarly works that categorize different aspects of speedrunning.
Besides the gain in public interest in speedrunning, only a small amount of such works on this field has emerged approximately in the last decade.
While a lot of these works focus on the interesting narrative and sociological consequences of the emergence of speedrunning
as a mode of play
\cite{NewmanPlayingVideogames08,Scully-BlakerPracticedPracticeSpeedrunning14,Scully-BlakerRecuratingAccidentSpeedrunning16,FordSpeedrunningTransgressiveplay18,Scully-BlakerSpeedrunningmuseumaccidents18,NewmanWrongWarpingSequence19,HayFullyOptimizedPost20,HemmingsenCodeLawSubversion20,RicksandTwereWellIt21},
this body also includes terminology and covers practical problems of speedrunning. 

Newman and Scully-Blaker both introduce nomenclature regarding the nature of speedruns and the involved activities employed by speedrunners.

\subsection{Newman's Activity Categories}
\label{subsec:newman}
In a recent work, Newman \cite{NewmanWrongWarpingSequence19} presents a thorough insight into the narrative consequences of speedrunning on the example of \emph{OoT}.
They also introduce categories of speedrunning activity: \emph{hidden affordances}, \emph{exploiting inconsistencies} and \emph{manipulation and reconstruction}. 

\begin{description}

\item[hidden affordances] In Newman's definition \emph{hidden affordances} can be broken down to optimally utilizing the designerly intended game mechanics.
For example in \emph{OoT} going backwards is faster than going forwards, so speedruns almost always consist of Link
(the protagonist of most of the games from the \emph{Legend of Zelda} franchise) walking backwards a lot of the time,
which is referred to as \emph{backwalking}\footnote{To put it in Newman's words: \q{That which is in front of Link is space already consumed.}\cite{NewmanWrongWarpingSequence19}}.
While Newman also states that \q{`designerly intent' is, obviously, difficult to assert}, they use as much official documentation about the game as possible to do so.

\item[exploiting inconsistencies] is the most expressive designation. Newman details it as \q{exploit[ing] programming bugs and
systemic errors in the code’s design and execution}\cite{NewmanWrongWarpingSequence19}. An example of this would be \emph{clipping}, a technique used in speedruns
of many games. \emph{Clipping} refers to the penetration of walls or other geometry in the game world the player is not supposed to go through. 

\item[manipulation and reconstruction] The third category of activity is denoted \emph{manipulation and reconstruction}. Newman describes it as follows:
\begin{quote}
    \q{The outcome of the technique is the creation of a connection in code that is
    subsequently rendered in polygonal space between regions in the game’s spatial and
    narrative architecture that are as palpably unintended by OoT ’s developers as they are
    disruptive and apparently injurious to the integrity of the game’s myth.} \cite{NewmanWrongWarpingSequence19}
\end{quote}
Activities of this category include techniques that consciously make use of the game's inner mechanics a player would normally be unaware of.
Creating techniques like these in many cases requires extensive knowledge about the game's logic and / or code.
The \emph{Reverse Bottle Adventure} (\emph{RBA}) technique for example makes use of knowledge about the IDs that each item in the game internally is identified by.
With this, executing very specific actions in the game can add or remove specific items from the player's inventory, based on the ID or amount of other items already
in the player's inventory \cite{ReverseBottleAdventure}.
\end{description}
These categories can help to define the earlier introduced term \emph{glitch}.
Although there is no universal consensus in the community, activities that fall into Newman's categories of \emph{exploiting inconsistencies} and
\emph{manipulation and reconstruction} are often referred to as \emph{glitches}. This work will follow this specification.

\subsection{Scully-Blaker's Speedrun Categories}
\label{subsec:scully-blaker}
In Scully-Blaker's work \cite{Scully-BlakerPracticedPracticeSpeedrunning14}, 
differentiations are made between \emph{finesse runs} and \emph{deconstructive runs}, regarding the degree of glitch 
exploitation and sequence breaking. 

\begin{description}
\item[finesse runs] are detailed as  \q{runs in which the player interacts 
with the game as an extreme extension of what a game designer 
may consider an `ideal' player [...] 
largely respecting the game's `narrative boundaries' while navigating them with an extreme level
of efficiency} \cite{Scully-BlakerPracticedPracticeSpeedrunning14}.
Put differently, one could say a \emph{finesse run} consists of playing a game as intended -- as far as ascertainable --
but very fast.

\item[deconstructive runs] on the other hand do not maintain any boundaries. These runs are \q{runs in which the
player exploits glitches within the game to break scripted sequences} \cite{Scully-BlakerPracticedPracticeSpeedrunning14},
skipping and reordering game content at will, as long as it serves the primary directive: speed.
\end{description}

\subsection{Routing}
\label{subsec:routing}

As outlined at the end of the \nameref{sec:introduction} (\ref{sec:introduction}),
speedruns have to be planned in advance, a procedure referred to as \emph{routing}.
Newman \cite{NewmanPlayingVideogames08} writes:
\begin{quote}
    \q{FPSs [First-Person-Shooters] are favoured because of their apparent non-linearity and the scope they seem to afford gamers
    to invent and create their own routes and develop their own styles that move them through the gameworld.}
\end{quote}
Newman goes on to also include Role-Playing-Games (RPGs) into these thoughts.

Speedrun routes consist of a number of events, connected by dependencies, starting with the
defined opening event and ending with one or multiple possible ending events. 
The problem of finding the ordering that takes the least amount of time possible is combinatorial in nature.
Some approaches to formalize and algorithmically address the problem of speedrun routing have been documented
\cite{Lafondcomplexityspeedrunningvideo18,IskovsTravellingmurdererproblem18,JstAnothrVirtuosoFindingOptimumNadeo19} 
but it is still underrepresented in research.
In these approaches, routing is usually defined as a graph shortest path problem.

Lafond \cite{Lafondcomplexityspeedrunningvideo18} takes the action-platform title \emph{Mega Man} as well as its successors \emph{Mega Man 2} and \emph{Mega Man 3}
as examples.
One key feature of these 2D Jump'n'Run variants is the aquisition of a new power after clearing each stage of the game.
These powers can be used to speed up subsequent stages.
All stages have to be cleared, while the order in which the stages are cleared is mostly left to the player.
Lafond defines sets of stages
$\mathbb S = \{S_1, ..., S_n\}$, with each stage $S_i = \{s_1, ..., s_k\}$
being a set of events on which time can be saved during a run\footnote{Symbols altered from original to prevent ambiguities}. 
This
save depends on previously completed stages, giving $s_j : \mathbb S \to \mathbb N$ as a function 
mapping each previously completed stage to a specific 
save for each event\footnotemark[\value{footnote}].
They form a dependency graph for a given game to be optimized, providing a route
with maximum time save through the game.
Even under favorable assumptions, Lafond \cite{Lafondcomplexityspeedrunningvideo18} proves this problem to be W[2]-hard \cite{downey2012parameterized}.

A more informal approach has been used by I\v skovs \cite{IskovsTravellingmurdererproblem18} with the working example of the popular game
\emph{The Elder Scrolls III: Morrowind}
As a typical open-world RPG title, there is a lot of traveling, character skill development and Non-Player-Character (NPC) relationships involved in the game mechanics.
The category I\v skovs takes on is \emph{all factions}, consisting of becoming the leader of all possible factions of NPCs in the game.
Taking into account the mentioned aspects among others, they formed an extensive \emph{quest dependency graph}.
I\v skovs then continues to use an evolutionary algorithm to generate and improve on possible routes through the game progression.
A lot of handcrafted customizations are then applied to minimize the run's time even further. This yields a very specific routing tool for a single game.
At the time of writing, there is only one entry in the corresponding leaderboard on speedrun.com, the creator of which
learned the route from I\v skovs' efforts \cite{VolvyRedditpostMorrowind18}.

Another informal example of algorithmic routing has been conducted by speedrunner JaV \cite{JstAnothrVirtuosoFindingOptimumNadeo19},
who breaks down a track in the car racing game \emph{TrackMania Nations Forever} to a variation of the travelling salesperson problem.
Track completion in this game is done by passing a number of checkpoints. In JaV's approach, these checkpoints are represented as a complete weighted digraph.
They then manually remove specific implausible edges from the graph and use a genetic algorithm
\cite{KirkFixedEndpointsOpen14} to generate near-optimal solutions. The used algorithm had to be adapted to avoid certain implausible combinations
and the solutions had to be curated, i.e. manually test-driven in the game to check for plausibility. Though this adaptions resulted in a specialized tool,
it is one of the rare examples that led to an improvement on the leaderboard times of a
game.

Taking it to the extreme, the process of routing can further be broken down to single simulation steps of a game's engine.
In 2009, a competition has been conducted with \q{The focus [...] on developing controllers that could play a version of Super Mario
Bros as well as possible} \cite{Togelius2009MarioAI10}.
The competition brought forth a number of submitted tools and algorithms to automate playing 
a variant of the popular game \emph{Super Mario World}.
The winner of the competition modeled the game state of each simulation step as a node of a graph,
and each possible next state (depending on different inputs) as adjacent nodes. 
Then, the shortest path to the goal -- or, as an approximation within available knowledge, to the right screen edge --
was determined through a variation of the A* algorithm \cite{HartFormalBasisHeuristic68}. 
The second and third best contributors of the competition used similar approaches.

\subsection{Criticism}
\label{subsec:criticism}
Glitches are often what differentiate Scully-Blaker's categories of runs from each other.
Scully-Blaker's definitions of run categories can be supported by Newmans terminology of activities.
While \emph{finesse runs} drive \emph{hidden affordances} to the extreme but leave the scripted game sequence and narrative structures mostly intact,
\emph{deconstructive runs} employ glitches and make heavy use of them, often
distorting and/or completely reassembling the event sequence.
However, these two categories are not to be seen as distinct.
The given descriptions are more likely two extremes of the space in which speedruns and their categories range.

Given this relation between Newman's and Scully-Blaker's terminology, connections to the leaderboard categories can be drawn as well.
As \emph{any\%} speedruns often seek to use whatever means possible to get to the defined end state, usage of glitches is prevalent.
More so, the used glitches can often be classified as Newman's \emph{exploiting inconsistencies} or \emph{manipulation and reconstruction} activities,
therefore these runs often range more in the realm of \emph{deconstructive runs}.
On the other side of the spectrum there are the \emph{Glitchless} leaderboard categories, which are governed by strict rules,
mostly oriented on narratively and designerly intentions.
As such, these runs can be categorized more in the realm of \emph{finesse runs}.

In a recent work, Ricksand \cite{RicksandTwereWellIt21} expresses disagreement with Scully-Blaker's categorizations,
listing lack of a definition of the term \emph{glitch}, lack of definition of completion requirements and inconsistencies in definition of an
ideal player as reasons.
However, considering distinctions already made in this paper, Scully-Blakers categories are considered suitable here.
As a slight addition, Scully-Blaker's definition of an ``ideal player'' can be extended by means of Newman's \emph{hidden affordances},
which then would include more techniques challenged by Ricksand like \emph{backwalking}.

Ricksand also questions the current procedure of rule acquisition in the speedrunning community. In their work, they suggest that the current process of voting on rulesets is flawed because it results in rules being arbitrary.
However, the argument can be made that speedrunning in its very essence is arbitrary.
Ultimately, Ricksand proposes the following model:
\begin{quotation}
    \q{Mechanic \textbf{m} in game \textbf{g} is allowed in a glitchless speedrun if and
    only if use of \textbf{m} does not contradict the fictional truth regarding the
    world in which the story of \textbf{g} takes place.} \cite{RicksandTwereWellIt21}
\end{quotation}
Not only does this not account for any technical glitch that doesn't get reflected by anything in the game world,
it also introduces the concept of fictional truth in order to legitimize glitchless rulesets.
Fictional truth however is partly up to the audience of a narrative \cite{CurrieFictionalTruth86}, rendering it arbitrary.

If nothing else, this disagreement shows the difficulty in deciding what should be categorized as a glitch, 
in scholarly work as well as in the speedrunning community.

The 2009 Mario AI Competition \cite{Togelius2009MarioAI10} did yield a number of good performing agents.
This competition can be seen as a TAS creation competition: 
Although there were no input prerecordings and the competition scoring incentivised total progression amount
rather than speed,
the submitted agents often played the game extremely fast.
Those efforts resemble TASs rather than RTA runs and have mostly been focusing on what will be called 
\emph{operational level routing} or \emph{operational routing} in this paper,
i.e. optimally planning or deciding on specific inputs at specific frames or simulation steps to reach a given short term goal or state.
This focus and the success of the developed tools is partly due to a property of the 2D-platformer nature of the \emph{Super Mario} titles --
going right on the screen is almost always a save indication of progress.
In fact, this metric has been used by the 2009 competition's winner 
as an heuristic for the A* algorithm \cite{Togelius2009MarioAI10}.
Optimizing the traversal of the game's progression mechanic meant going right as quickly as possible.
However, this holds true only for the specific variation of the \emph{Super Mario} title used 
(consisting only of horizontally oriented levels, progressing to the right)
and only if isolated levels are taken into account.
In contrast, to minimize the total time of an entire game's completion as the previously presented works did,
\emph{strategic level routing} or \emph{strategic routing} is employed.
As a result, speedruns of games which rely heavily on \emph{strategic routing} could not yet benefit from these advances.
The strategic layer of planning the traversal between levels
-- or, more generally speaking, between different events, from a speedrun's opening event to an ending event --
is generally what is referred to as \emph{routing} in the speedrunning community as it is in this work.

In many cases, routing is not a one-time process. Especially with \emph{deconstructive runs} \cite{Scully-BlakerPracticedPracticeSpeedrunning14}
with a high amount of \emph{inconsistency exploiting} and \emph{manipulation and reconstruction} activities, 
routes are significantly reworked with new glitches and exploits being found,
due to their often disruptive nature regarding the game's intended event sequence.
This act of altering the game's intended order of progression by means of glitches is referred to as
\emph{sequence breaking} in the speedrunning community.
This leads to a challenge when defining the underlying routing problem,
as these often unpredictable changes in routing possibilities have to be accounted for.
\emph{Glitchless} speedruns can mostly be characterized as \emph{finesse runs}
\cite{Scully-BlakerPracticedPracticeSpeedrunning14},
as the use of glitches is prohibited in these rulesets.
Therefore, routing underlies more consistent structures and can easier be defined formally.

Some challenging aspects of speedrunning have been engaged by the presented works, others are yet left unaccounted for.

%% file: content/approaches.tex
\section{Envisioned Models and Challenges}
\label{sec:approaches}

With the knowledge compiled to this point, the actual speedrun routing optimization problem can be approached.
To do so, this problem has to be formalized and defined as a model.
Given the outlined nature of the routing problem,
graph representations will be favored.
The presented models are considered visions that can be elaborated and improved on rather than final solutions.
Again, \emph{OoT} will be used as the working example and
results may be less applicable to other games or genres.

\subsection{Weighted Game Event Digraph}
\label{subsec:WGD}

As a first approach, a weighted game digraph is assumed
\begin{equation}
    \label{eq:gamegraph}
    G = (V, E, w) \, ,
\end{equation}
with nodes $V = \{v_1, ..., v_n\}$ as the set of all events relevant to the game's progression.
Each possible traversal between these events make up the set of directed edges $E = \{e_1, ..., e_m\}$ between the nodes representing the given events.
Edges are weighted with a function $w: E \to \mathbb R$, assigning each edge the time it takes to traverse between the in-game events in the given direction.
Edge directions are considered as this traversal time can differ by direction.
Note that the term traversal is used as opposed to travel, as node traversal does not necessarily involve in-game movement. 
The routing process would then mean to find the shortest path between node $v_\alpha$ resembling the starting event and $v_\omega$ representing the ending event -- or a set $V_\omega$ of ending events --, 
traversing all events $V_r \subseteq V$ deemed required by the category's ruleset.

This rudimentary model already introduces several challenges:
\begin{enumerate}
    \item \emph{Defining the Nodes}: ``Events relevant to the game's progression'' is not defined and it is nontrivial to do so.
    \item \emph{Defining the Edges}: The extent of ``each possible traversal between events'' has underlying restrictions that have to be defined as well.
    \item \emph{Dynamic Weights}: Edge weights are not consistent but rather change with graph traversal.
    \item \emph{Repeatable Events}: A subset of the events can be repeated once triggered, while others can not, further increasing the complexity.
    \item \emph{Multiobjective Optimization}: The model does not account for any dimension other than time in a possible route.
\end{enumerate}

These challenges will be elaborated on in succession, followed by possible ways to accommodate them within the problem model.

\subsubsection{Defining the Nodes}
\label{subsubsec:nodes}

As for the definition of ``events relevant to the game's progression'', in the spirit of optimization it could be argued that only the items
needed to qualify for the category should be involved, i.e. reducing \eqref{eq:gamegraph} to
\begin{equation}
    \label{eq:reducedgamegraph}
    G_r = (V_r, E_r, w_r) \, ,
\end{equation}
with $E_r \subseteq E$ being the reduced set of edges remaining between $V_r$ -- weighted by time of traversal $w_r$ -- ,
as everything else would constitute a detour and thus cost additional overall time.
Routing could be accomplished by finding the shortest topological ordering of $G_r$.
However, an optimization frequently employed by speedrunners renders this reduction unsuitable.

\emph{OoT} speedruns make heavy use of the resources at the player's disposal, including but not limited to Link's health, explosives (bombs and bombchus)
and (in-game) time of day. A detour early on in a route to get more explosives can significantly speed up the rest of the route, as many travelling times can be
drastically reduced due to glitches using bombs as a key component. For example, the bombchu item -- a moving, explosive device -- is not required by many \emph{OoT} categories, 
yet many speedruns start out their route by collecting the nearest supply of bombchus in order to speed up the rest of the route enough to make it worthwhile. Thus, in fact, the opposite argument could be made: Expanding the set of events by all possible
item pickups and refill locations as well as taking resource management into account can minimize the time even further.
This obviously would increase the problem's complexity.

\subsubsection{Defining the Edges}
\label{subsubsec:edges}

The definition of ``each possible traversal between events'' depends on the focus of the route.
Considering only RTA viable routes this can be put as ``humanly doable and complying to the category's ruleset''.
Finding all possible edges and assigning weights to them would require manually checking for category viability and timing the traversal between
any two events in the game in both directions, which would be very time consuming and error prone.
To counteract this complexity increase, nodes can be clustered by spatial proximity.
Events and items that are very close to each other and / or obtainable by similar means can be clustered to single nodes.
Another measure can be to approximate traversal times or using duration categories as integer numbers instead of actual timings, trading reduced complexity for  uncertainties.

\subsubsection{Dynamic Weights}
\label{subsubsec:weights}
As suggested above, edge weights are not static, but can and frequently do change with the already traversed nodes and in some cases
(e.g. when in-game time of day is involved) with time.
Even when considering only required events $V_r$, the items collected can also enhance the player's ability to traverse the game world in a speedy
fashion in different ways they may or may not be intended for.
This effectively changes the edge weights $w$ while traversing the game graph.
This \emph{hidden gain} can lead to faster overall routes by taking detours early on that may have been undiscovered when only taking the starting weights into account.

Lafond \cite{Lafondcomplexityspeedrunningvideo18} partly accommodated this by weighting the graph edges differently.
For weighting an edge from stage $S_i$ to $S_j$ they use the time saved in $S_j$ as the result of doing $S_i$ beforehand,
rather than the time needed to traverse between the stages.
Considering multiple previously cleared stages and thus possibly multiple ways of clearing a stage, it is assumed that for every event the
immediate best alternative is used.

The \emph{OoT} equivalent for this would be to weight an edge from event $v_i$ to $v_j$ by the time saved by traversing $v_i$ before $v_j$.
This could constitute a way to significantly simplify the game graph, as many items and events do not provide any time save and thus have no impact on traversal speed.

Another way to handle dynamic edge weights would be to model game states rather than events as the graph's nodes.
Depending on the information considered for defining the states, this can lead to a very large amount of nodes.
This is elaborated on in section \emph{\nameref{subsec:stategraph}} (\ref{subsec:stategraph}).

\subsubsection{Repeatable Events}
\label{subsubsec:repeat}

Some of the events mentioned above are repeatable, especially when considering item pickups and refills.
For example, grass patches in \emph{OoT} can drop bomb pickups when destroyed, and the patch regrows every time the location is reloaded.
Thus, to repeat and benefit from a desired event,
it might be required to traverse to another node and back to the desired event's node.
Other events are one-time only.
This holds true for all pickups and items from chests as well as for most story progression events.
These events' nodes can be removed from consideration once traversed.
Section \emph{\nameref{subsec:stategraph}} (\ref{subsec:stategraph}) discusses the possible implications of this on a graph representation.

\subsubsection{Multiobjective Optimization}
\label{subsubsec:multi}

The objective of speedrunning this far has only been considered a one-dimensional problem, solely dictated by time.
While speed is in fact the primary directive of speedrunning, there is another dimension that can be taken into account: difficulty.
Many speedrunning techniques require very precise inputs, sometimes precise to the frame.
Some techniques aren't even possible to be performed by humans but are exclusive to TASs.
Similarly, other aspects can be taken into consideration when constructing a route, like
the aforementioned \emph{hidden gain} or resource management.
A consideration of possible changes to a graph model is given in section \emph{\nameref{subsec:vector}} (\ref{subsec:vector}).

\subsection{Weighted Game State Graph}
\label{subsec:stategraph}

One way to engage the challenge of \emph{\nameref{subsubsec:weights}} (\ref{subsubsec:weights})
is to consider the nodes $V_s$ of a state graph $G_s$, reflecting game states, rather than events:
\begin{equation}
    \label{eq:stategraph}
    G_s = (V_s, E_s, w_s)
\end{equation}
Similar to a \emph{\nameref{subsec:WGD}} (\ref{subsec:WGD}) from \eqref{eq:gamegraph},
the edges $E_s$ represent traversals between different states of the game,
weighted by the time taken to do so $w_s$.
If these states account for all necessary elements described above (location, items, item counts, time of day, etc.)
this would eliminate edge weight dynamics.
However, this would lead to an immense increase of nodes and edges.
In fact, the presence of repeatable events would likely render this graph infinite.

\subsection{Vector Valued Game Graph}
\label{subsec:vector}

The challenge of \emph{\nameref{subsubsec:multi}} (\ref{subsubsec:multi}) can be taken into account by a graph
\begin{equation}
    \label{eq:vectorgamegraph}
    G_m = (V, E, \mathbf{w}_m) \, ,
\end{equation}
with a number $n$ of weighting dimensions, which would yield vector valued edge weights and thus a
vector valued weighting function $\mathbf{w}_m : E_m \to \mathbb R^n$.
For example, a second vector element can be used to assign difficulty ratings to graph traversals.
This can be used to parameterize possible routing algorithms by the runner's skill level.
When including TASs, very high difficulty ratings can be assigned to TAS-only traversals.

Another possible application for vector valued edge weights is to represent the \emph{hidden gain} introduced in \emph{\nameref{subsubsec:weights}} (\ref{subsubsec:weights}) as a vector element. 
This would constitute an approximation, but could be a useful reduction of complexity. Weight dynamics could be replaced by this.
In case of repeatable events, this \emph{hidden gain} value might decrease when the corresponding node or edge is traversed,
to prevent indefinite looping.
Also, difficulty and \emph{hidden gain} ratings can be combined to yield three-dimensional vectors.

Vector valued edge weights open up the problem to multiobjective optimization \cite{miet99,ehrg05} and pathfinding algorithms
\cite{StewartMultiobjective91,Rajabi-BahaabadiMultiobjectivepathfinding15}, 
resulting in non-dominated sets of possible routes, possibly further increasing creativity and idea sparking in speedrun routing.

The events $V$ and their traversals $E$ can also be replaced by states $V_s$ and their respective traversals $E_s$ from \eqref{eq:stategraph},
resulting in a combination with a \emph{\nameref{subsec:stategraph}} (\ref{subsec:stategraph}),
introducing all advantages and disadvantages pointed out there.

%% file: content/algorithm.tex
\section{Prospective Solutions}
\label{sec:algorithm}

After a model has been chosen to represent the routing problem, an optimization method has to be found.
All concepts outlined are again specific to \emph{OoT}, applicability to other games and genres has to be assessed.
Good solutions should consider the challenges identified in \emph{\nameref{sec:approaches}} (\ref{sec:approaches})
and look out for new, unidentified challenges.
Of course, solutions should also respect the complexity of a given run. For example, a \emph{glitchless} route for a mostly linear game with \emph{finesse run} characteristics resembles different challenges than an \emph{any\%} route for a highly non-linear game.

When modeling an \emph{Event Graph} with traversal times as edge weights, 
conventional pathfinding algorithms like Dijkstra's \cite{DijkstraNoteTwoProblems59} or A* \cite{HartFormalBasisHeuristic68} 
are unsuitable given the edge dynamics.
Same applies to algorithms like Multiobjective A* \cite{StewartMultiobjective91} in the case of vector valued edge weights.
For edge weights to become static, a \emph{State Graph} with extensively detailed states is needed, vastly increasing the amount of nodes.

Considering time save as edge weight similar to Lafond's approach \cite{Lafondcomplexityspeedrunningvideo18} can simplify the graph.
However, some edge dynamics will still prevail, as some techniques need multiple items to be collected or events to be triggered beforehand.

Two of the presented works on speedrun routing \cite{JstAnothrVirtuosoFindingOptimumNadeo19,IskovsTravellingmurdererproblem18}
employed evolutionary algorithms (EAs) \cite{Baec97,eibe15,jong16} to handle the complexity involved.
An approach using EAs for (multiobjective) optimization \cite{deb01,coel07} seems promising, including assertions to exclude implausible combinations.
Next to the implementation of different EA approaches, also the use of other metaheuristics is envisioned \cite{gendreau2010handbook,Soer15}. For example, ant colony optimization algorithms \cite{DorigoAntcolonyoptimization99} have already proved their suitability to solve pathfinding problems \cite{RishiwalFindingOptimalPaths10,Mocholiemotionallybiasedant10}.

Another possible approach is to conduct the traversal by an agent, deciding further progression after every edge traversal.
Decision making can be conducted by above mentioned approaches, combinations of them, or by other means, 
such as a neuronal network taking information about the graph and previous traversal as input.
Applications of machine learning for decision making in game agents -- especially in form of deep reinforcement learning --
suggest potential in this approach \cite{shengyi21,lample16,Ye_2020}.
With regard to edge dynamics, an agent must take Link's inventory, 
the previously traversed nodes and edges as well as other relevant information into account
to maintain the Markov property.

%% file: content/conclusions.tex
\section{Conclusions and Outlook}
\label{sec:conclusions}

Routing is an integral part of most speedrun categories, regardless of the game.
As such, having tools at disposal to automate this process can greatly help speedrunners and support speedrunning very complex games and genres. 
In addition, such tools are expected to be transferable to complex problems from logistics, huge vehicle routing instances from globally operating companies for example. 
However, as was shown in this work as well as the given references, speedrun routing is hard, whether done manually or algorithmically.

Given the complexity of \emph{OoT}'s routing setting and the fact that despite its age new,
heavily exploitable glitches and techniques are being found to this day,
it is unreasonable to assume that a route will ever be found that cannot be improved on in the future.
In fact this can be said for most games with enough combinatorial complexity to its progression mechanics.
In these cases, routing algorithms can be seen as a tool for speedrunners to support their routing efforts by
new influences.
Moreover, there are speedrunners who enjoy manual routing as an important part of speedrunning.
Solving this problem computationally would, if possible at all, render this process obsolete.

This work provides a first overview of works important for the subject area of speedrunning. 
These are critically discussed and relevant terminology is pointed out to support professional discussion.
Moreover, different approaches are presented to model the identified speedrunning routing optimization problem and corresponding challenges are depicted.
Finally, prospective solutions for the challenges are outlined.

Future works can discuss further on one or more of the outlined approaches for modeling and solving the problem of speedrun routing.
Applying the findings to other games and genres can uncover further possibilities not apparent when only focusing on a single game.

As a final note, the topic suggests itself for competitions.
Participants can compete in different ways, such as finding the most accurate route representation model,
solving given representations, or of course creating a full route through a given game.
In the latter case, support of actual speedrunners would be needed to assess plausibility.

%% file: bibliography/ludography.tex
\section*{Ludography}
\label{sec:ludography}

{
\footnotesize

\begin{description}
\item[The Legend of Zelda: Ocarina of Time (OoT) {[Nintendo 64]}]
Developer / Publisher: Nintendo, Japan  (1998)
\item[Super Mario World {[SNES]}]
Developer / Publisher: Nintendo, Japan (1990)
\item[Mega Man/2/3 {[NES]}]
Developer / Publisher: Capcom, Japan (1987/1988/1990 resp.)
\item[The Elder Scrolls III: Morrowind {[PC (Windows), Xbox]}]
Developer: Bethesda Softworks (2002),
Publisher: Ubisoft, France
\item[TrackMania Nations Forever {[PC (Windows)]}]
Developer: Nadeo (2008),
Publisher: Focus, France; Enlight, USA; Deep Silver, Germany; Digital Jesters, UK
\end{description}

}

%% file: main.bbl
\begin{thebibliography}{10}
\providecommand{\url}[1]{\texttt{#1}}
\providecommand{\urlprefix}{URL }
\providecommand{\doi}[1]{https://doi.org/#1}

\bibitem{GDQTrackerEvent}
{{GDQ Tracker}} - {{Event List}},
  \url{https://gamesdonequick.com/tracker/events/}, [accessed 14. May 2021]

\bibitem{GDQStatus}
{GDQStat.us}, \url{https://gdqstat.us/previous-events/agdq-2020/?series=0},
  [accessed 13. May 2021]

\bibitem{ReverseBottleAdventure}
Reverse {{Bottle Adventure}} - {{ZeldaSpeedRuns}},
  \url{https://www.zeldaspeedruns.com/oot/ba/reverse-bottle-adventure},
  [accessed 14. May 2021]

\bibitem{speedruncom}
speedrun.com, \url{https://www.speedrun.com/oot}, [accessed 17. May 2021]

\bibitem{Baec97}
{Back}, T., {Hammel}, U., {Schwefel}, H.P.: Evolutionary computation: comments
  on the history and current state. IEEE Transactions on Evolutionary
  Computation  \textbf{1}(1),  3--17 (1997). \doi{10.1109/4235.585888}

\bibitem{coel07}
{Coello Coello}, C.A., {Lamont}, G.B., {van Veldhuizen}, D.A.: {Applications Of
  Multi-Objective Evolutionary Algorithms }. World Scientific, New Jersey, 2.
  edn. (2007)

\bibitem{CurrieFictionalTruth86}
Currie, G.: Fictional {{Truth}}. Philosophical Studies: An International
  Journal for Philosophy in the Analytic Tradition  \textbf{50}(2),  195--212
  (Sep 1986). \doi{10.1007/BF00354588}

\bibitem{jong16}
De~Jong, K.A.: Evolutionary Computation: A Unified Approach. MIT Press,
  Cambridge, MA (2016)

\bibitem{deb01}
Deb, K.: {Multi-Objective Optimization Using Evolutionary Algorithms}. Wiley,
  Chichester, UK (2001)

\bibitem{DijkstraNoteTwoProblems59}
Dijkstra, E.W.: {A Note on Two Problems in Connexion with Graphs.} Numerische
  Mathematik  \textbf{1},  269--271 (1959)

\bibitem{DorigoAntcolonyoptimization99}
Dorigo, M., Di~Caro, G.: Ant colony optimization: A new meta-heuristic. In:
  Proceedings of the 1999 {{Congress}} on {{Evolutionary
  Computation}}-{{CEC99}} ({{Cat}}. {{No}}. {{99TH8406}}). vol.~2, pp.
  1470--1477 Vol. 2 (Jul 1999). \doi{10.1109/CEC.1999.782657}

\bibitem{downey2012parameterized}
Downey, R.G., Fellows, M.R.: Parameterized complexity. Springer Science \&
  Business Media, New York (1999)

\bibitem{ehrg05}
Ehrgott, M.: {Multicriteria Optimization}. Springer, Berlin, 2nd edn. (2005)

\bibitem{eibe15}
Eiben, A.E., Smith, J.E.: {Introduction to Evolutionary Computing}. Natural
  Computing Series, Springer, Berlin, 2. edn. (2015)

\bibitem{FordSpeedrunningTransgressiveplay18}
Ford, D.: {Speedrunning: Transgressive play in digital space}. In: {Proceedings
  of Nordic DiGRA 2018}. {Bergen, Norway} (Nov 2018).
  \doi{10.13140/RG.2.2.12357.91369}

\bibitem{gendreau2010handbook}
Gendreau, M., Potvin, J.Y., et~al.: Handbook of metaheuristics, vol.~3.
  Springer, Cham, Switzerland (2019)

\bibitem{HartFormalBasisHeuristic68}
Hart, P.E., Nilsson, N.J., Raphael, B.: A {{Formal Basis}} for the {{Heuristic
  Determination}} of {{Minimum Cost Paths}}. IEEE Transactions on Systems
  Science and Cybernetics  \textbf{4}(2),  100--107 (Jul 1968).
  \doi{10.1109/TSSC.1968.300136}

\bibitem{HayFullyOptimizedPost20}
Hay, J.: Fully {{Optimized}}: {{The}} ({{Post}})human {{Art}} of
  {{Speedrunning}}. Journal of Posthuman Studies  \textbf{4}(1),  5--24 (2020).
  \doi{10.5325/jpoststud.4.1.0005}

\bibitem{HemmingsenCodeLawSubversion20}
Hemmingsen, M.: Code is {{Law}}: {{Subversion}} and {{Collective Knowledge}} in
  the {{Ethos}} of {{Video Game Speedrunning}}. Sport, Ethics and Philosophy
  pp. 1--26 (Jul 2020). \doi{10.1080/17511321.2020.1796773}

\bibitem{shengyi21}
Huang, S., Bamford, C., Ontanon, S., Grela, L.: {Gym-µRTS: Toward Affordable
  Full Game Real-time Strategy Games Research with Deep Reinforcement
  Learning}. In: {{IEEE Conference}} on {{Games}} (05 2021).
  \doi{10.13140/RG.2.2.18639.82081}

\bibitem{IskovsTravellingmurdererproblem18}
I{\v s}kovs, A.: Travelling murderer problem: Planning a {{Morrowind}}
  all-faction speedrun with simulated annealing (Apr 2018),
  \url{https://www.kimonote.com/@mildbyte/travelling-murderer-problem-planning-a-morrowind-all-faction-speedrun-with-simulated-annealing-part-1-41079/},
  [accessed 13. May 2021]

\bibitem{JstAnothrVirtuosoFindingOptimumNadeo19}
{JstAnothrVirtuoso}: Finding the {{Optimum Nadeo Cut}}... {{With Science}}!!
  (May 2019), \url{https://www.youtube.com/watch?v=1ZsAjvO9E1g}, [accessed 13.
  May 2021]

\bibitem{KirkFixedEndpointsOpen14}
Kirk, J.: Fixed {{Endpoints Open Traveling Salesman Problem}} - {{Genetic
  Algorithm}} (May 2014),
  \url{https://www.mathworks.com/matlabcentral/fileexchange/21197-fixed-endpoints-open-traveling-salesman-problem-genetic-algorithm},
  [accessed 14. May 2021]

\bibitem{Lafondcomplexityspeedrunningvideo18}
Lafond, M.: The complexity of speedrunning video games. In: Ito, H., Leonardi,
  S., Pagli, L., Prencipe, G. (eds.) 9th {{International Conference}} on
  {{Fun}} with {{Algorithms}} ({{FUN}} 2018). Leibniz {{International
  Proceedings}} in {{Informatics}} ({{LIPIcs}}), vol.~100, pp. 27:1--27:19.
  {Schloss Dagstuhl\textendash Leibniz-Zentrum fuer Informatik}, {Dagstuhl,
  Germany} (2018). \doi{10.4230/LIPIcs.FUN.2018.27}

\bibitem{lample16}
Lample, G., Chaplot, D.S.: Playing {FPS} games with deep reinforcement
  learning. CoRR  (2016), \url{http://arxiv.org/abs/1609.05521}

\bibitem{miet99}
Miettinen, K.: {Nonlinear Multiobjective Optimization}. Kluwer, Boston, MA
  (1999)

\bibitem{Mocholiemotionallybiasedant10}
Mocholi, J.A., Jaen, J., Catala, A., Navarro, E.: An emotionally biased ant
  colony algorithm for pathfinding in games. Expert Systems with Applications
  \textbf{37}(7),  4921--4927 (Jul 2010). \doi{10.1016/j.eswa.2009.12.023}

\bibitem{NewmanPlayingVideogames08}
Newman, J.: Playing with {{Videogames}}. {Routledge}, {London} (Jun 2008)

\bibitem{NewmanWrongWarpingSequence19}
Newman, J.: Wrong {{Warping}}, {{Sequence Breaking}}, and {{Running}} through
  {{Code}}. Journal of the Japanese Association for Digital Humanities
  \textbf{4}(1),  7--36 (2019). \doi{10.17928/jjadh.4.1\_7}

\bibitem{Rajabi-BahaabadiMultiobjectivepathfinding15}
{Rajabi-Bahaabadi}, M., {Shariat-Mohaymany}, A., Babaei, M., Ahn, C.W.:
  Multi-objective path finding in stochastic time-dependent road networks using
  non-dominated sorting genetic algorithm. Expert Systems with Applications
  \textbf{42}(12),  5056--5064 (Jul 2015). \doi{10.1016/j.eswa.2015.02.046}

\bibitem{RicksandTwereWellIt21}
Ricksand, M.: ``{{Twere Well It Were Done Quickly}}'': {{What Belongs}} in a
  {{Glitchless Speedrun}}? Game Studies  \textbf{21}(1) (Apr 2021),
  \url{http://gamestudies.org/2101/articles/ricksand}, [accessed 26. May 2021]

\bibitem{RishiwalFindingOptimalPaths10}
Rishiwal, V., Yadav, M., Arya, K.V.: Finding {{Optimal Paths}} on {{Terrain
  Maps}} using {{Ant Colony Algorithm}}. International Journal of Computer
  Theory and Engineering  \textbf{2}(3),  416--419 (Jan 2010).
  \doi{10.7763/IJCTE.2010.V2.178}

\bibitem{Scully-BlakerPracticedPracticeSpeedrunning14}
{Scully-Blaker}, R.: A {{Practiced Practice}}: {{Speedrunning Through Space
  With}} de {{Certeau}} and {{Virilio}}. Game Studies  \textbf{14}(1) (Aug
  2014), \url{http://gamestudies.org/1401/articles/scullyblaker}, [accessed 26.
  May 2021]

\bibitem{Scully-BlakerRecuratingAccidentSpeedrunning16}
{Scully-Blaker}, R.: Re-Curating the {{Accident}}: {{Speedrunning}} as
  {{Community}} and {{Practice}}. Masters, Concordia University (Sep 2016)

\bibitem{Scully-BlakerSpeedrunningmuseumaccidents18}
{Scully-Blaker}, R.: {The Speedrunning museum of accidents}. Kinephanos
  (Preserving Play, Special Issue),  71--88 (Aug 2018),
  \url{https://www.kinephanos.ca/2018/the-speedrunning-museum-of-accidents/},
  [accessed 26. May 2021]

\bibitem{StewartMultiobjective91}
Stewart, B.S., White, C.C.: Multiobjective {{A}}*. Journal of the ACM
  \textbf{38}(4),  775--814 (Oct 1991). \doi{10.1145/115234.115368}

\bibitem{Szita2012}
Szita, I.: Reinforcement Learning in Games, pp. 539--577. Springer Berlin
  Heidelberg, Berlin, Heidelberg (2012). \doi{10.1007/978-3-642-27645-3\_17}

\bibitem{Soer15}
Sörensen, K.: Metaheuristics—the metaphor exposed. International
  Transactions in Operational Research  \textbf{22}(1),  3--18 (2015).
  \doi{10.1111/itor.12001}

\bibitem{Togelius2009MarioAI10}
Togelius, J., Karakovskiy, S., Baumgarten, R.: The 2009 {{Mario AI
  Competition}}. In: {{IEEE Congress}} on {{Evolutionary Computation}}.
  pp.~1--8 (Jul 2010). \doi{10.1109/CEC.2010.5586133}

\bibitem{VolvyRedditpostMorrowind18}
Volvy: Reddit post about the {{Morrowind}} all factions speedrun route (Nov
  2018),
  \url{www.reddit.com/r/speedrun/comments/9u1r9o/using_ai_to_grind_out_routes/e91dg6w/},
  [accessed 31. May 2021]

\bibitem{Ye_2020}
Ye, D., Liu, Z., Sun, M., Shi, B., Zhao, P., Wu, H., Yu, H., Yang, S., Wu, X.,
  Guo, Q., Chen, Q., Yin, Y., Zhang, H., Shi, T., Wang, L., Fu, Q., Yang, W.,
  Huang, L.: Mastering complex control in moba games with deep reinforcement
  learning. Proceedings of the AAAI Conference on Artificial Intelligence
  \textbf{34}(04),  6672--6679 (Apr 2020). \doi{10.1609/aaai.v34i04.6144}

\end{thebibliography}
